\documentclass{article}
\usepackage{spconf,amsmath,amssymb,amsfonts,graphicx,hyperref}
\usepackage{booktabs}
\usepackage{url}

\title{Deep Attention-guided Adaptive Subsampling}
\name{Sharath M Shankaranarayana, Soumava Kumar Roy, Prasad Sudhakar, Chandan Aladahalli}
\address{GE Healthcare, Bangalore, India}
\begin{document}
%
\maketitle
\begin{abstract}
Although deep neural networks have provided impressive gains in performance, these improvements often come at the cost of increased computational complexity and expense. In many cases, such as 3D volume or video classification tasks, not all slices or frames are necessary due to inherent redundancies. To address this issue, we propose a novel learnable subsampling framework that can be integrated into any neural network architecture.
Subsampling, being a non-differentiable operation, poses significant challenges for direct adaptation into deep learning models. While some works, have proposed solutions using the Gumbel-max trick to overcome the problem of non-differentiability, they fall short in a crucial aspect: they are only task-adaptive and not input-adaptive. Once the sampling mechanism is learned, it remains static and does not adjust to different inputs, making it unsuitable for real-world applications.
To this end, we propose an attention-guided sampling module that adapts to inputs even during inference. This dynamic adaptation results in performance gains and reduces complexity in deep neural network models. We demonstrate the effectiveness of our method on 3D medical imaging datasets from MedMNIST3D as well as two ultrasound video datasets for classification tasks, one of them being a challenging in-house dataset collected under real-world clinical conditions.

\end{abstract}
\begin{keywords}
Learnable Sampling, Deep Learning, 3D Classification, Video Classification
\end{keywords}
%


\section{Introduction}

The proliferation of high-dimensional data, such as videos and volumetric scans, has underscored the critical need for efficient processing methods that balance computational cost and informational fidelity. Sampling or sub-sampling data frames or slices of interest helps in situations where exhaustive processing of all data frames/slices is infeasible. Conventional techniques often employ uniform sampling or handcrafted heuristics, which are frequently suboptimal as they fail to discern and prioritize the most salient information within the data. 
Recent developments in machine learning have led to the paradigm of learnable sampling, wherein  the most informative data components are identified in a data-driven manner. Deep Probabilistic Subsampling (DPS) \cite{huijben2020deep} was a seminal work in this domain, proposing a differentiable framework for learning task-adaptive sampling patterns. While effective, DPS learns a fixed, content-agnostic strategy. Active Deep Probabilistic Subsampling (ADPS) \cite{van2021active} extended this by enabling instance adaptive sampling, where the selection is contingent previously sampled components.

Despite these advances, existing methods face notable limitations. First, the static nature of DPS's sampling patterns, while optimal on average, may be ill-suited for individual data instances. Second, although ADPS introduces adaptiveness at the instance level, it does so by conditioning only on the already sampled components, without explicitly leveraging the input features themselves. This limits its ability to fully exploit the content-specific structure of the data during inference. To overcome these challenges, we propose Deep Attention-guided Subsampling (DAS), a novel framework that synergizes differentiable sampling with attention mechanisms for dynamic and adaptive selection of informative data slices, frames, or features from volumetric or sequential data. Our approach leverages the Gumbel-Softmax reparameterization trick \cite{jang2017categorical, maddison2017concrete} to ensure end-to-end differentiability for discrete sample selection. The core innovation of DAS is its attention-driven, input adaptive sampling strategy.

The primary contributions of this work are:
\begin{itemize}
    \item A novel, plug-and-play neural sampling module for dynamic sampling in 3D volumes and videos, which adapts to the input at inference time, rendering the approach both task and input adaptive.
    \item A demonstration of the proposed framework's efficacy on classification tasks across eight medical imaging datasets, including six from MedMNIST3D \cite{medmnist3d}, one public ultrasound video dataset \cite{busv}, and a proprietary in-house dataset collected in a clinical setting.
\end{itemize}
\begin{figure}[!tp]
\centering
\includegraphics[width=0.98\linewidth]{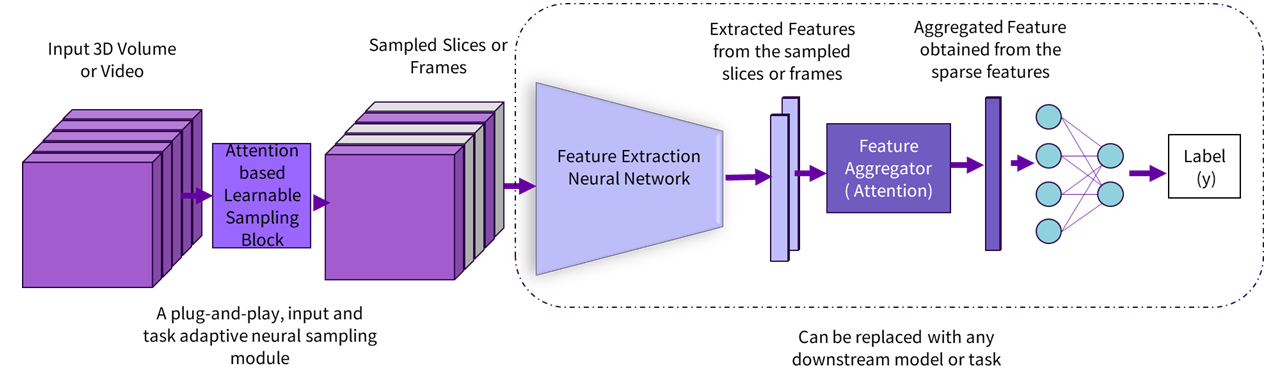}
\caption{Our proposed Sampling Module can be plugged into any downstream task}
\label{fig:vid_vol_classification}
\end{figure}
While this paper focuses on classification of 3D medical volumes and videos, the proposed framework is broadly applicable to various downstream tasks and is designed for efficient inference in resource-constrained environments. By intelligently selecting only the most informative data components, DAS achieves substantial computational savings while maintaining or improving performance compared to other sampling methods. Another key feature of the approach is that the sampling matrix itself is produced as an output, enabling explicit control and interpretability of the sampling process.

\subsection{Related Work}
\label{sec:related_work}
Our work builds upon recent advancements in learnable subsampling, attention mechanisms, and differentiable sampling techniques for efficient video and volumetric data processing.
The concept of learning sampling patterns for computational efficiency was notably advanced by Deep Probabilistic Subsampling (DPS) \cite{huijben2020deep}, which introduced a differentiable framework to learn a fixed, task-adaptive sampling strategy. While effective, its content-agnostic nature limits its applicability to inputs requiring instance-specific analysis. Active Deep Probabilistic Subsampling (ADPS) \cite{van2021active} addressed this by enabling content-adaptive sampling, but its multi-stage process can introduce significant computational overhead at inference time, undermining the goal of efficiency. Our proposed method advances this line of work by introducing a lightweight, single-pass module that performs input-adaptive sampling dynamically during inference. Attention mechanisms have been widely used to identify salient frames in videos \cite{korbar2019scsampler, yeung2016end}. However, these methods often lack end-to-end differentiability or are not integrated within a unified sampling framework. The development of differentiable sampling techniques, particularly the Gumbel-Softmax trick \cite{jang2017categorical, maddison2017concrete}, has been pivotal in enabling the training of networks that make discrete selections. Our work synergizes these two areas, using an attention mechanism to guide a Gumbel-Softmax-based sampler. This allows our framework to be both highly adaptive and end-to-end trainable, focusing computational resources on the most informative data slices in a principled manner. By doing so, we create a lightweight and efficient neural sampling module that remains adaptive at inference time, addressing the key limitations of prior approaches.
\section{Proposed Method}
\label{sec:methods}

\begin{figure}[!tp]
\centering
\includegraphics[width=0.98\linewidth]{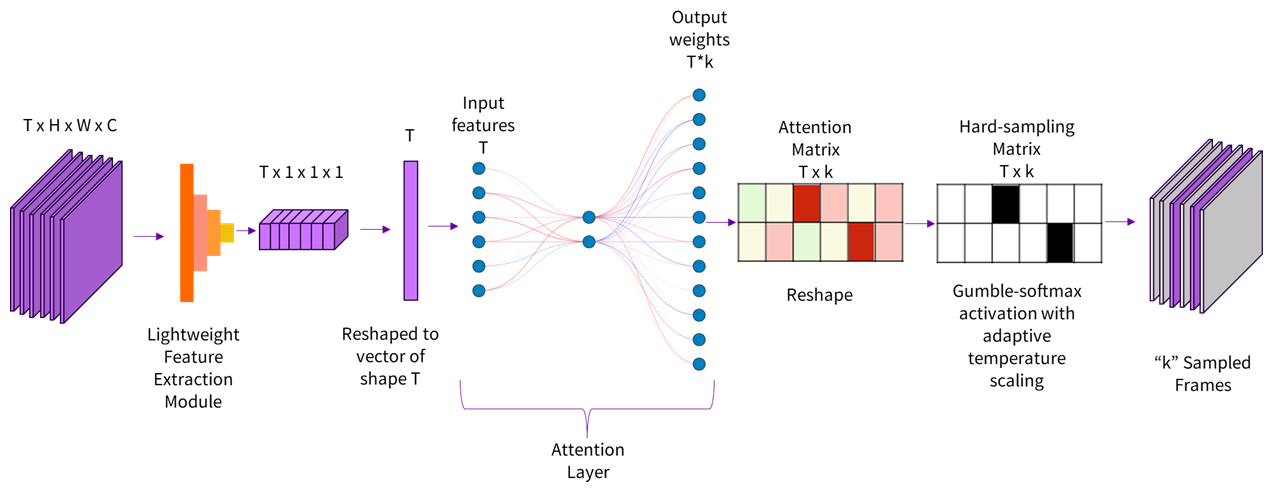}
\caption{An overview of the proposed deep attention-guided subsampling framework. The model takes a sequence of frames, extracts features, and uses an attention mechanism to generate a sampling matrix. This matrix is then used to select a subset of frames.}
\label{fig:architecture}
\end{figure}

Our work, Deep Attention-guided Subsampling (DAS), introduces a framework for task- and input-adaptive subsampling of medical video sequences (or 3D medical image volumes). Although, as mentioned before, the proposed framework can be applied to any task, however, in this paper we focus on 3D volume and video classification tasks. Given a sequence of $T$ frames $\mathbf{X} \in \mathbb{R}^{B \times T \times C \times H \times W}$, our objective is to learn a sampling function $\mathcal{S}_{\boldsymbol{\theta}}$ that selects a subset of $k$ frames ($k \ll T$). The overall architecture, depicted in Figure \ref{fig:architecture}, comprises a lightweight feature extraction module, an attention layer for generating sampling probabilities, and a Gumbel-Softmax sampling mechanism.

\subsection{Lightweight Feature Extraction}
To make the sampling input adaptive, we first extract a feature vector using a lightweight feature extraction module. This module consists of multiple parallel pathways to compute a rich representation of the input sequence. The extracted features are aggregated into a tensor $\mathbf{F} \in \mathbb{R}^{B \times T \times d}$, where $d$ is the feature dimension. This comprehensive representation captures both the content of individual frames and their relationships within the sequence. The modalities are detailed below.

To capture temporal dynamics, we compute the frame-wise variance over the spatial and channel dimensions. 



To identify anatomical boundaries, we apply a set of complementary Sobel and Laplacian kernels to each frame to compute edge magnitudes. This allows the model to focus on frames with significant structural information. 
The extracted features are then concatenated to form a comprehensive feature representation $\mathbf{F}$

\subsection{Multi-Head Attention Layer}
The aggregated feature tensor $\mathbf{F}$ is processed by a multi-head attention layer to generate the final sampling logits. This mechanism allows the model to jointly attend to information from different representation subspaces. Our implementation employs $H$ parallel attention heads, enabling the capture of diverse sampling patterns.

The core of our attention mechanism is the dynamic generation of head-specific scale factors, which modulate a shared base attention distribution. Let $\mathbf{a}_{\text{base}} \in \mathbb{R}^{B \times T}$ be the base attention scores, derived from one or more of the feature modalities (e.g., temporal variance). For each attention head $h \in \{1, \dots, H\}$, a dedicated Multi-Layer Perceptron, $\text{MLP}_h$, generates a vector of $k$ scale factors, $\mathbf{s}_h \in \mathbb{R}^{B \times k}$, from the comprehensive feature representation $\mathbf{F}$:
\begin{equation}
\mathbf{s}_h = \text{Softplus}(\text{MLP}_h(\mathbf{F}))
\end{equation}
The Softplus activation ensures that the scaling factors are non-negative.

Each head then computes its own attention matrix, $\mathbf{A}_h \in \mathbb{R}^{B \times k \times T}$, by scaling the base attention scores. For the $j$-th sample to be selected (where $j \in \{1, \dots, k\}$), the attention scores across all $T$ frames are modulated by the $j$-th component of the scale factor vector $\mathbf{s}_h$:
\begin{equation}
\mathbf{A}_h^{(:,j,:)} = \mathbf{a}_{\text{base}} \odot \mathbf{s}_h^{(:,j)}
\end{equation}
where $\odot$ denotes element-wise multiplication with broadcasting.

The final attention matrix $\mathbf{A} \in \mathbb{R}^{B \times k \times T}$ is obtained by averaging the outputs of the individual heads, effectively creating an ensemble of specialized samplers:
\begin{equation}
\mathbf{A} = \frac{1}{H} \sum_{h=1}^{H} \mathbf{A}_h
\end{equation}
Each of the $k$ rows in this final attention matrix corresponds to a single frame to be sampled, and the values across the $T$ columns represent the unnormalized probabilities, or logits, of selecting each input frame for that particular sample.
\subsection{Differentiable Gumbel-Softmax Sampling}
To allow for end-to-end training, we employ the Gumbel-Softmax trick for differentiable sampling from the categorical distribution defined by the attention logits.

We propose \textit{adaptive temperature scaling}, wherein the temperature parameter $\tau$ of the Gumbel-Softmax distribution is adaptively scaled based on the input features. This allows the model to dynamically control the trade-off between exploration (high temperature) and exploitation (low temperature) during training. The temperature is modulated as follows:
\begin{equation}
\tau = \tau_0 \cdot \left(0.5 + \sigma\left(\text{MLP}_{\text{temp}}(\mathbf{F})\right)\right)
\end{equation}
where $\tau_0$ is a base temperature, $\sigma$ is the sigmoid function, and $\text{MLP}_{\text{temp}}$ is a small neural network. This formulation keeps $\tau$ within a stable range of $[0.5\tau_0, 1.5\tau_0]$, preventing extreme values while allowing for controlled adaptation.

For each of the $k$ frames to be sampled, we first obtain a "soft" probability distribution over the $T$ input frames by applying the Gumbel-Softmax function to the corresponding row of the attention matrix $\mathbf{A}$:
\begin{align}
    \mathbf{G}_{b,j,t} &\sim \text{Gumbel}(0,1) \\
    \mathbf{P}_{\text{soft}} &= \text{Softmax}_t\left(\frac{\mathbf{A} + \mathbf{G}}{\tau}\right)
\end{align}
where $\mathbf{P}_{\text{soft}} \in \mathbb{R}^{B \times k \times T}$ is the matrix of soft probabilities. For the forward pass, we use a hard, one-hot version of these probabilities, $\mathbf{P}_{\text{hard}}$, obtained by taking the $\arg\max$ over the soft probabilities. To ensure differentiability, we use a straight-through estimator (STE) as proposed in \cite{jang2017categorical}.
The resulting matrix $\mathbf{P} \in \mathbb{R}^{B \times k \times T}$ is our final sampling matrix. The sampled frames (or slices) are obtained by applying this matrix to the input sequence. 


\section{Experiments and Results}
\begin{table*}[!tp]
\centering
\caption{Classification performance on public datasets. We report both AUC and Accuracy (\%) for comprehensive evaluation. Best subsampling method per dataset and metric is \textbf{bolded}.}
\label{tab:results_public}
\resizebox{\textwidth}{!}{%
\begin{tabular}{lcccccccccccc}
\toprule
& \multicolumn{2}{c}{\textbf{Full Sequence}} & \multicolumn{2}{c}{\textbf{Random}} & \multicolumn{2}{c}{\textbf{Uniform}} & \multicolumn{2}{c}{\textbf{DPS}} & \multicolumn{2}{c}{\textbf{ADPS}} & \multicolumn{2}{c}{\textbf{DAS (Ours)}} \\
\cmidrule(lr){2-3} \cmidrule(lr){4-5} \cmidrule(lr){6-7} \cmidrule(lr){8-9} \cmidrule(lr){10-11} \cmidrule(lr){12-13}
\textbf{Dataset} & \textbf{AUC} & \textbf{Acc} & \textbf{AUC} & \textbf{Acc} & \textbf{AUC} & \textbf{Acc} & \textbf{AUC} & \textbf{Acc} & \textbf{AUC} & \textbf{Acc} & \textbf{AUC} & \textbf{Acc} \\
\midrule
Organ & 0.943 & 63.9 & 0.916 & 55.5 & 0.927 & 55.9 & 0.912 & 57.0 & 0.928 & 57.3 & 0.931 & 58.1 \\
Nodule & 0.802 & 80.6 & 0.770 & 72.9 & 0.779 & 74.1 & 0.776 & 74.5 & 0.779 & 0.782 &0.799 & 75.8\\
Adrenal & 0.671 & 76.4 & 0.543 & 67.1 & 0.569 & 59.8 & 0.590 & 63.4 & 0.591 & 67.6 & 0.597 & 75.1 \\
Fracture & 0.572 & 40.0 & 0.521 & 34.6 & 0.502 & 34.9 & 0.513 & 34.1 & 0.581 & 42.1 & 0.584 & 42.1\\
Vessel & 0.788 & 86.7 & 0.724 & 78.5 & 0.714 & 79.2 & 0.636 & 71.7 & 0.739 & 80.7 & 0.752 & 82.9  \\
Synapse & 0.649 & 71.0 & 0.555 & 58.8 & 0.565 & 63.9 & 0.546 & 60.5 & 0.613 & 64.7 & 0.628 & 70.7 \\
\midrule
BVUS & 0.711 & 72.9 & 0.679 & 62.1 & 0.708 & 70.2 & 0.714 & 70.2 & 0.753 & 74.9 & 0.740 & 75.6 \\
\bottomrule
\end{tabular}%
}
\end{table*}
\label{sec:experiments}
We validate our proposed Deep Attention-guided Subsampling (DAS) framework through comprehensive experiments across diverse medical imaging datasets, comparing against established baselines and state-of-the-art learned subsampling methods.

We evaluate DAS on eight medical imaging datasets spanning different modalities and clinical tasks:

\textbf{MedMNIST3D\cite{medmnist3d}:} Six 3D volumetric datasets (Organ, Nodule, Adrenal, Fracture, Vessel, Synapse) covering multi-organ segmentation and pathology detection tasks. Each 3D volume is treated as a sequence of 2D slices for consistent evaluation.
\textbf{Breast Ultrasound Video (BUSV) \cite{busv}:} A public ultrasound video dataset for breast lesion detection, providing a representative real-world video classification (binary) benchmark.
\textbf{In-house Gastric Antrum Dataset:} A proprietary clinical ultrasound video dataset collected in a real-world hospital setting which comprises five-class gastric content classification.

\subsection{Experimental Setup}
All methods are evaluated using an identical downstream architecture for fair comparison:

\textbf{Feature Extraction:} MobileNetV3-Small \cite{howard2019searching}, pretrained on ImageNet and fine-tuned per dataset, extracts frame/slice-level features. This lightweight backbone ensures computational efficiency while maintaining representational capacity.

\textbf{Temporal Aggregation:} A learnable attention mechanism aggregates the feature sequence into a unified representation, as illustrated in Fig.~\ref{fig:vid_vol_classification}.

\textbf{Classification:} A two-layer MLP maps the aggregated features to class predictions.

We evaluate classification tasks using balanced accuracy and AUC metrics. All results are averaged over three independent runs. Models are trained using Adam optimizer (lr=1e-4, batch size=16) with early stopping on validation loss.

We compare against several subsampling strategies: \textbf{Full Sequence}, where all frames or slices are processed, representing the computational upper bound; \textbf{Random Sampling}, which selects $k$ frames uniformly at random; \textbf{Uniform Sampling}, which selects equidistant frames across the sequence; \textbf{Deep Probabilistic Subsampling (DPS)} \cite{huijben2020deep}, a task-adaptive but content-agnostic learned sampling method, \textbf{Active Deep Probabilistic Subsampling (ADPS)} \cite{van2021active}, which is input-adaptive but solely by conditioning on previously sampled frames rather than directly on the input itself.; and \textbf{DAS (Ours)}, which is both input-adaptive and task-adaptive. For all subsampling methods we select $50\%$ of the original sequence length. For both ultrasound video datasets, we normalize the input sequences to have a uniform number of frames to facilitate batch training and inference. This is achieved by padding each video with empty frames at the end, based on the maximum sequence length observed in the dataset.
\begin{table}[!tp]
\centering
\caption{Performance on in-house Gastric Antrum dataset. We report both AUC and Accuracy (\%) metrics. DAS achieves better than full-sequence performance with significantly reduced computational cost.}
\label{tab:results_inhouse}
\begin{tabular}{lcc}
\toprule
\textbf{Method} & \textbf{AUC} & \textbf{Accuracy (\%)} \\
\midrule
Full Sequence (Baseline) & 0.611 & 30.1 \\
Random Sampling & 0.564 & 26.4 \\
Uniform Sampling & 0.639 & 29.0 \\
DPS & 0.479 & 11.1 \\
ADPS & 0.582 & 27.5 \\ 
DAS (Ours) & \textbf{0.639} & \textbf{34.1} \\
\bottomrule
\end{tabular}
\end{table}
\subsection{Results}

Our results consistently demonstrate the superiority of DAS across all evaluated datasets. On public benchmarks (Table \ref{tab:results_public}), DAS generally outperforms both DPS and ADPS, with the exception of the \textit{Synapse} dataset. It should be noted that both ADPS and DAS reach performance close to the full-sequence baseline on many of the datasets indicating redundancy in the data for classification task which can be exploited by superior sampling strategies. On the \textit{BVUS} dataset, ADPS performs better in terms of AUC, achieving $0.711$ compared to DAS's $0.639$, whereas DAS outperforms ADPS in accuracy, with $75.6\%$ versus $74.9\%$. The data-adaptive/learnable sampling strategies perform better than the baseline in case of \textit{BVUS} dataset because of the inherent redundancy in the ultrasound videos and also because of the padding. Notably, on the challenging in-house Gastric Antrum dataset (Table \ref{tab:results_inhouse}), DAS significantly surpasses even the full-sequence baseline. This is particularly impactful given the nature of real-world clinical ultrasound scans, which are often noisy due to factors such as intermittent probe contact—resulting in empty frames—and high gain settings. These findings highlight the effectiveness of input-adaptive subsampling for deployment in resource-constrained medical imaging environments.


\section{Conclusion}
\label{sec:conclusion}

In this work, we introduced the Deep Attention-guided Subsampling (DAS) framework, a novel approach for efficient and adaptive subsampling in medical imaging tasks. By leveraging an attention-guided mechanism, DAS dynamically adapts to input data even during inference, enabling both input-adaptive and task-adaptive sampling. Our method demonstrated significant performance improvements across diverse datasets, including challenging real-world clinical scenarios, while reducing computational complexity.
A promising direction for future work is the development of a learnable feature extraction module that can automatically identify salient features for guiding the sampling process,further enhancing the performance of DAS.

\bibliographystyle{IEEEbib}
\bibliography{refs}

\end{document}